\documentclass[sigconf]{acmart}
\pdfoutput=1

\usepackage{booktabs} 
\usepackage{bm}
\usepackage{subcaption}
\usepackage{mathrsfs}
\usepackage[ruled]{algorithm2e}
\usepackage{amsmath}
\usepackage{caption}
\usepackage{enumitem}
\usepackage{comment}
\usepackage{graphicx}
\usepackage{amsthm}
\usepackage{multirow}
\usepackage{color}
\usepackage{dsfont}

\newtheorem*{conjecture*}{Conjecture}
\newtheoremstyle{nonindented}{1ex}{1ex}{}{}{\bfseries}{.}{.5em}{}
\newtheoremstyle{indented}{1ex}{1ex}{\itshape\addtolength{\leftskip}{0.6cm}\addtolength{\rightskip}{0.6cm}}{}{\bfseries}{.}{.5em}{}
\theoremstyle{nonindented}
\theoremstyle{indented}
\theoremstyle{plain}

\newcommand{\union}{\cup}

\renewcommand{\hat}{\widehat}

\def\min{\qopname\relax n{min}}
\def\max{\qopname\relax n{max}}

\newcommand{\eat}[1]{}

\newenvironment{lp*}{\begin{equation*}  \begin{array}{lll}}{\end{array}\end{equation*}}

\usepackage{tikz-qtree}
\usepackage{tikz-qtree-compat}
\usetikzlibrary{shapes,decorations,arrows,calc,arrows.meta,fit,positioning}
\tikzset{
	-Latex,auto,node distance =1 cm and 1 cm,semithick,
	state/.style ={ellipse, draw, minimum width = 0.7 cm},
	point/.style = {circle, draw, inner sep=0.04cm,fill,node contents={}},
	bidirected/.style={Latex-Latex,dashed},
	el/.style = {inner sep=2pt, align=left, sloped}
}

\setcopyright{acmcopyright}
\copyrightyear{2023}
\acmYear{2023}
\setcopyright{rightsretained}
\acmConference[KDD '23] {Proceedings of the 29th ACM SIGKDD Conference on Knowledge Discovery and Data Mining}{August 6--10, 2023}{Long Beach, CA, USA.}
\acmBooktitle{Proceedings of the 29th ACM SIGKDD Conference on Knowledge Discovery and Data Mining (KDD '23), August 6--10, 2023, Long Beach, CA, USA}
\acmPrice{}
\acmISBN{979-8-4007-0103-0/23/08}
\acmDOI{10.1145/3580305.3599408}

\AtBeginDocument{%
 }

\copyrightyear{2023}
\acmYear{2023}
\setcopyright{rightsretained}
\acmConference[KDD '23]{Proceedings of the 29th ACM SIGKDD Conference on Knowledge Discovery and Data Mining}{August 6--10, 2023}{Long Beach, CA, USA}
\acmBooktitle{Proceedings of the 29th ACM SIGKDD Conference on Knowledge Discovery and Data Mining (KDD '23), August 6--10, 2023, Long Beach, CA, USA}
\acmDOI{10.1145/3580305.3599408}
\acmISBN{979-8-4007-0103-0/23/08}

\makeatletter
\gdef\@copyrightpermission{
  \begin{minipage}{0.3\columnwidth}
   \href{https://creativecommons.org/licenses/by/4.0/}{\includegraphics[width=0.90\textwidth]{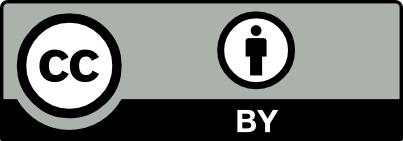}}
  \end{minipage}\hfill
  \begin{minipage}{0.7\columnwidth}
   \href{https://creativecommons.org/licenses/by/4.0/}{This work is licensed under a Creative Commons Attribution International 4.0 License.}
  \end{minipage}
  \vspace{5pt}
}
\makeatother

\begin{document}
\title{Learning for Counterfactual Fairness from Observational Data}

\author{Jing Ma}
\affiliation{%
\institution{University of Virginia}
\city{}
\state{}
\country{}
}
\email{jm3mr@virginia.edu}

\author{Ruocheng Guo}
\affiliation{%
\institution{Bytedance Research}
\city{}
\state{}
\country{}
}
\email{ruocheng.guo@bytedance.com}

\author{Aidong Zhang}
\affiliation{%
\institution{University of Virginia}
\city{}
\state{}
\country{}
}
\email{aidong@virginia.edu}

\author{Jundong Li}
\affiliation{%
\institution{University of Virginia}
\city{}
\state{}
\country{}
}
\email{jundong@virginia.edu}

\renewcommand{\shortauthors}{Jing Ma, Ruocheng Guo, Aidong Zhang, \& Jundong Li}

\newcommand{\mymodel}{CLAIRE}
\newcommand{\mymodela}{CLAIRE-M}
\newcommand{\mymodelb}{CLAIRE-A}
\newcommand{\cfpa}{CFP-U}
\newcommand{\cfpb}{CFP-O}
\newcommand{\bigCI}{\mathrel{\text{\scalebox{1.07}{$\perp\mkern-10mu\perp$}}}}
\newtheorem{assumption}{Assumption}

%

%

\begin{abstract}
Fairness-aware machine learning has attracted a surge of attention in many domains, such as online advertising, personalized  recommendation, and social media analysis in web applications. Fairness-aware machine learning aims to eliminate biases of learning models against certain subgroups described by certain protected (sensitive) attributes such as race, gender, and age. Among many existing fairness notions, counterfactual fairness is a popular notion defined from a causal perspective. It measures the fairness of a predictor by comparing the prediction of each individual in the original world and that in the counterfactual worlds in which the value of the sensitive attribute is modified. A prerequisite for existing methods to achieve counterfactual fairness is the prior human knowledge of the causal model for the data. However, in real-world scenarios, the underlying causal model is often unknown, and acquiring such human knowledge could be very difficult. In these scenarios, it is risky to directly trust the causal models obtained from information sources with unknown reliability and even causal discovery methods, as incorrect causal models can consequently bring biases to the predictor and lead to unfair predictions. In this work, we address the problem of counterfactually fair prediction from observational data without given causal models by proposing a novel framework \mymodel. Specifically, under certain general assumptions, \mymodel~ effectively mitigates the biases from the sensitive attribute with a representation learning framework based on counterfactual data augmentation and an invariant penalty. Experiments conducted on both synthetic and real-world datasets validate the superiority of \mymodel~ in both counterfactual fairness and prediction performance.  

\end{abstract}

%



\maketitle

\section{Introduction}
Recent years have witnessed a rapid development of machine learning based prediction \cite{schwartz2004fair,corbett2018measure,brennan2009evaluating} in various high-impact applications such as personalized recommendation \cite{wu2021learning,mehrotra2018towards}, ranking in searches \cite{geyik2019fairness,pitoura2020fairness}, and social media analysis \cite{leonelli2021fair,aguirre2021gender}. 
Recent literatures \cite{berk2018fairness} have shown that the predictions based on traditional machine learning often exhibit biases against certain demographic subgroups that are described by certain protected attributes (a.k.a. sensitive attributes) such as race, gender, age, and sexual orientation. 
Thus, how to develop a \textit{fair} predictor has attracted a surge of attentions \cite{grgic2016case,hardt2016equality,zafar2017fairness,zemel2013learning,bellamy2019ai,bird2019fairness,wadsworth2018achieving}. Among them, the seminal work of \textit{counterfactual fairness} \cite{kusner2017counterfactual} makes use of the causal mechanism to model how discrimination is exhibited, and eliminates it at the individual level based on the Pearl's causal structural models \cite{pearl2009causal}. The intuition of counterfactual fairness is to encourage the predictions made from different versions of the same individual to be equal. {
For example, the predictions for 
``in an online talent search, how would a certain candidate be ranked if this candidate had been a male/female?"
should be identical to achieve the notion of counterfactual fairness.}

A prerequisite of existing methods to achieve counterfactual fairness is the prior human knowledge of causal models.
A causal model \cite{pearl2009causal,pearl2009causality} typically consists of a causal graph and the corresponding structural equations that describe the causal relationships among different variables.
Existing works on counterfactual fairness \cite{kusner2017counterfactual,russell2017worlds,wu2019counterfactual,xu2019achieving} overwhelmingly rely on the assumption that the underlying causal model is (at least partially) known and correct, in order to mitigate the biases across different sensitive subgroups.
However, existing work often suffers from the following major limitation: In real world, the underlying causal model is often unknown, especially when the data is high-dimensional \cite{belloni2017program,wang2020high}. The construction of a trustworthy causal model often requires knowledge from domain experts, which is expensive in both time and labor. In addition, it is extremely challenging to validate the correctness of the obtained causal model. Without external guidance of human knowledge, other existing works mostly rely on causal discovery techniques \cite{spirtes2000causation,pearl2009causality,kalisch2007estimating,le2016fast,heckerman1999bayesian,spirtes2016causal} to learn the causal model from observational data, but these methods can suffer from various mistakes in discovering the causal relations, and thus lead the predictor to pick up biased information of the sensitive attribute \cite{nauta2019causal}. 

\begin{figure}[t]
\centering
        \includegraphics[width=.47\textwidth]{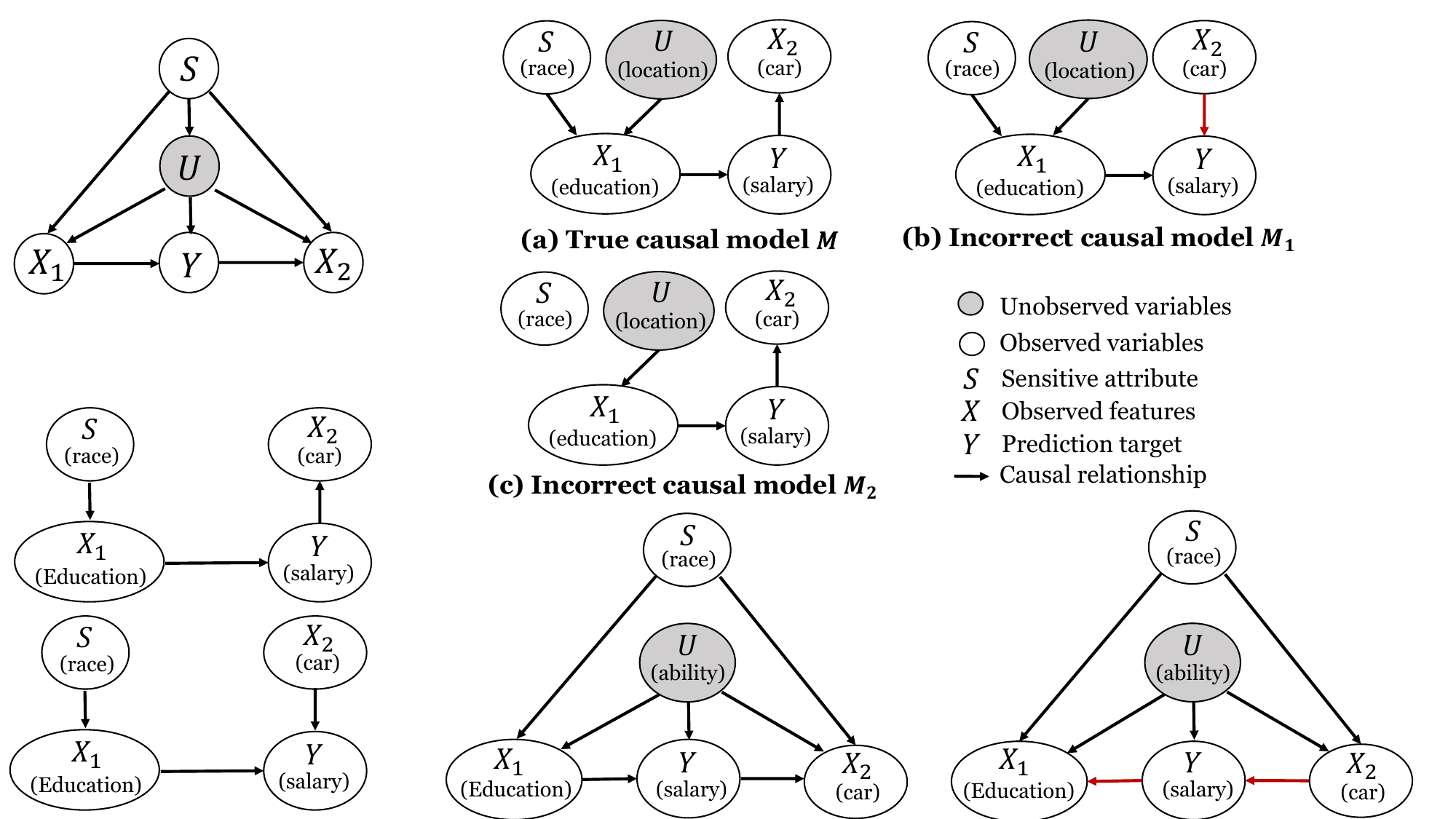}
        \caption{An illustrative example of incorrect causal models.} 
         \label{fig:example1}
 \end{figure}

Here, the toy example in Fig.~\ref{fig:example1} intuitively explains two scenarios with incorrect causal models.
Fig.~\ref{fig:example1}(a) shows an example of a true causal model (often determined by domain experts) in which we aim to predict the salary (prediction target $Y$) of people in different races (described by the sensitive attribute $S$). We assume that the level of education (observed feature $X_1$) of each person is a cause, and the salary also influences the type of car each person would like to purchase (observed feature $X_2$). Unobserved variables $U$ (e.g., geographic location) could also have a causal effect on the observed variables. 
To learn a counterfactually fair predictor, most existing works \cite{kusner2017counterfactual,russell2017worlds} utilize a given causal model, and only use those variables which are not causally influenced by the sensitive attribute (i.e., non-descendants of $S$) for prediction. 
We now consider two cases when the given causal model is incorrect: 
1) Consider an incorrect causal model $\mathcal{M}_1$ in Fig.~\ref{fig:example1}(b), where the direction of the causal relation $Y\rightarrow X_2$ is reversed (highlighted in red). Note that $X_2$ is causally influenced by $S$ in the true causal model $\mathcal{M}$. If a predictor is based on $\mathcal{M}_1$, $X_2$ would be directly used in prediction, and thus it violates counterfactual fairness with biases from the sensitive attribute. 
%
2) Consider another incorrect causal model $M_2$, where an existing causal relation $S\rightarrow X_1$ in the true causal model $\mathcal{M}$ is ignored. Predictors based on $M_2$ would directly use $X_1$ in prediction, which results in biases. Unfortunately, causal models are quite common to be incorrectly assumed or discovered  \cite{spirtes2000causation,pearl2009causality,kalisch2007estimating,le2016fast}.
To address the aforementioned issues of insufficient human knowledge of causal model, we study a novel problem of \textit{learning counterfactually fair predictor with unknown causal models}. 
Although it is in principle impossible to achieve counterfactual fairness without any causal model \cite{kusner2017counterfactual}, we take initial explorations to mitigate the unfairness based on certain general assumptions, and circumvent the prerequisite of explicit prior knowledge. 
%
However, this studied problem remains a daunting task mainly due to the following challenges: 
1) In order to achieve counterfactual fairness, the causal effect from the sensitive attribute $S$ to the prediction must be removed \cite{kusner2017counterfactual,russell2017worlds}, but an unknown causal model brings challenges to track the influence of the sensitive attribute and eliminate the biases; 
2) There might exist unobserved variables which can be used to predict the target (e.g., ``geographic location" in the salary prediction example), 
but without a correct causal model, it is harder to capture these unobserved variables for prediction due to the lack of prior knowledge regarding these variables. 
3) 
Many factors (e.g.,  failure in obtaining correct causal relations) 
may lead to unfair predictions, but it is difficult to exclude their influence without a correct causal model. 
In a nutshell, all of these challenges are essential due to the lack of counterfactual data. 


To tackle these challenges, we propose a novel framework --- \textit{ \textbf{C}ounterfactua\textbf{L}ly f\textbf{AI}r and invariant p\textbf{RE}dictor} \textit{(\mymodel)}, which learns counterfactually fair representations for target prediction. 
%
To remove the biases from sensitive attributes without any given causal model (challenge 1), we develop a counterfactual data augmentation module to implicitly capture the causal relations in data, and generate counterfactuals for each individual with different sensitive attribute values. 
In this way, \mymodel~ can learn fair representations by using a counterfactual fairness constraint to minimize the difference between the predictions made on the original data and on its counterfactuals.
To capture the unobserved variables which can help counterfactually fair prediction (challenge 2), 
\mymodel~ maps the observed variables to a latent representation space to encode the unobserved variables that can facilitate the prediction. The aforementioned counterfactual fairness constraint can preserve those unobserved variables which are not biased. 
%
%
To further reduce the factors which potentially impede counterfactual fairness (challenge 3), we exclude the variables with \textit{spurious correlations} to the target (i.e., variables that appear to be causal to the target but are not, e.g., $X_2$ in Fig.~\ref{fig:example1}(a)) from the learned representations. Spurious correlations can easily lead to incorrect causal models. Besides, removing these variables can often benefit model prediction performance, as shown in \cite{arjovsky2019invariant}. 
We summarize our main contributions as follows: 
\begin{itemize}
    \item  \textbf{Problem:} We study an important problem of learning counterfactually fair predictor from observational data. We analyze its importance, challenges, and impacts. 
\item  \textbf{Algorithm:} We propose a novel framework \mymodel~ for this problem. Specifically, we learn  fair representations based on counterfactual data augmentation. Besides, we exclude spurious correlations to further reduce potential biases. 
\item  \textbf{Experiments:} We conduct extensive experiments to evaluate our framework on synthetic and real-world datasets. The results show that \mymodel~ outperforms the existing baselines.
\end{itemize}

\section{Preliminaries}

\subsection{Notations}
In this paper, we use upper-cased letters, e.g., $X$, to denote random variables, lower-cased letters, e.g., $x$, to denote specific values.
$P(X)$ refers to the probabilistic function of $X$. We use $X$, $S$, $U$, $Y$ to represent the observed non-sensitive features/attributes, sensitive attribute, unobserved variables, prediction label/target for any instance, respectively. 
Specifically, we use $X^s,Y^s$ to denote the corresponding features and target of any instance with the observation of a specific sensitive attribute value $S=s$, where $s\in \mathcal{S}$, and $\mathcal{S}$ is the space of the sensitive attribute value. 
$\hat{Y}$ denotes the predicted label (for classification tasks) or target (for regression tasks).

\subsection{Counterfactual Fairness}
Counterfactual fairness \cite{kusner2017counterfactual} is an individual-level fairness notion based on the causal mechanism. 
It is built upon the Pearl's causal framework \cite{pearl2009causal}, which is defined as a triple $(U, V, F)$ such that: 
\begin{itemize}
    \item $U$ is the set of latent variables, which are often assumed to be exogenous and consequently independent of each other;
    \item $V$ is a set of observed variables, which are endogenous and determined by variables in $U\union V$;
    \item $F=\{f_1(\cdot),f_2(\cdot),...,f_{|V|}(\cdot)\}$ is a set of functions (referred to as \textit{structural equations}) which describe the causal relationships among the above variables. For each variable $V_i\in V$, $ V_i = f_i(pa_i,U_{pa_i})$, where ``$pa_i\subseteq V \setminus{V_i}$” and ``$U_{pa_i}\subseteq U$" are variables that directly determine $V_i$. 
\end{itemize}
A causal model is associated with a \textit{causal graph}, which is a directed acyclic graph (DAG). Each node in the causal graph corresponds to a variable in the causal model, and each directed edge represents a causal relationship. For example, for observed variables $A, B$, the value of the \textit{counterfactual} "what would $A$ have been if $B$ had been set to $b$?" is denoted by $A_{B\leftarrow b}$. 

Based on a given causal model, a predictor uses a function $\hat{Y}=f(X,S)$ to make the prediction for each instance. The predictor is \textit{counterfactually fair} \cite{kusner2017counterfactual} if under any context $X=x$ and $S=s$, 
\begin{equation}
P(\hat{Y}_{S\leftarrow s}=y|X=x,S=s) = P(\hat{Y}_{S\leftarrow s'}=y|X=x,S=s),
\end{equation}
for all $y$ and $s'\ne s$. Here $\hat{Y}_{S\leftarrow s}=f(X_{S\leftarrow s},s)$ denotes the prediction made on the counterfactuals when the value of $S$ had been set to $s$.



\subsection{Biases under Incorrect Causal Models}
To achieve the notion of counterfactual fairness, existing works often~\cite{kusner2017counterfactual,russell2017worlds} follow a two-step process: 1) First, they use the observed data to fit the causal model and infer the posterior distribution $P(U|X,S)$ of unobserved variables $U$; 2) Second, they train a counterfactually fair predictor based on the fitted causal model. In particular, this step can be achieved in different ways: an initial work \cite{kusner2017counterfactual} trains the predictor with only unobserved variables $U$ and the non-descendants of $S$ as input. We refer to this method as \cfpa. Another work \cite{russell2017worlds} considers a counterfactual fairness objective $|f(X_{S\leftarrow s},s)-f(X_{S\leftarrow s'},s')|$ for each instance, aiming to minimize the difference between the predictions made on different counterfactuals of the sensitive attribute. We refer to this method as \cfpb. In this subsection, we use some simple examples to show the biases in the prediction of these existing counterfactual fairness methods when the given causal model is incorrect.

\noindent\textbf{Example 1. } First, we consider the case when the counterfactual fairness methods have been given an incorrect causal model as shown in Fig.~\ref{fig:example1}(b).  
In the aforementioned salary prediction example,
the ground truth causal model $\mathcal{M}$ is shown in Fig. \ref{fig:example1}(a). It indicates that people's salary can causally influence their choices of cars to purchase. 
In this example, we let the causal model $\mathcal{M}$ be as follows:
$$
P(S=1)=0.5, P(S=0)=0.5, \epsilon_1,\epsilon_y,\epsilon_2 \sim \mathcal{N}(0,1),
$$
$$
X_1 \leftarrow S + U + \epsilon_1, Y \leftarrow X_1 + \epsilon_y, X_2 \leftarrow Y + \epsilon_2.
$$
$X_2$ is correlated with $Y$ because it is $Y$'s child node, but this correlation may lead the model to incorrectly take $X_2$ as one of $Y$'s parent nodes, as the incorrect causal model $\mathcal{M}_1$ shown in Fig. \ref{fig:example1}(b). Then the goal of counterfactual fairness:
$
P(\hat{Y}_{S\leftarrow s}^{\mathcal{M}_1}|X=x,S=s) = P(\hat{Y}_{S\leftarrow s'}^{\mathcal{M}_1}|X=x,S=s)
$ defined on $\mathcal{M}_1$ is different from what is defined on the true causal model $\mathcal{M}$.
Based on the incorrect causal model $\mathcal{M}_1$, \cfpa~ will take $X_2$ as an input to the predictor, but $X_2$ contains biased information because it is actually a descendant of the sensitive attribute, thus it will bring bias into prediction. For \cfpb, if we assume a linear predictor $\hat{Y}=W_1 X_1 + W_2 X_2 + W_S S$, then the fairness penalty on the incorrect causal model would be:
$$
\mathbb{E}(|f(X^{\mathcal{M}_1}_{S\leftarrow 1},1)-f(X^{\mathcal{M}_1}_{S\leftarrow 0},0)|) = |W_1+W_S|,
$$
while the fairness penalty based on the true causal model would be:
$$
\mathbb{E}(|f(X^{\mathcal{M}}_{S\leftarrow 1},1)-f(X^{\mathcal{M}}_{S\leftarrow 0},0)|) = |W_1+W_2+W_S|.
$$
Such difference can lead to inappropriate learning results for the parameters in the predictor. As the fairness penalty based on the incorrect causal model has no constraint on $W_2$, the predictor can not exclude the biases contained in $X_2$.

\noindent\textbf{Example 2. } 
We now consider another case of incorrect causal model shown in Fig.~\ref{fig:example1}(c). 
In the salary prediction example, consider that the dataset contains a majority
sensitive subgroup $S=0$ (e.g., race A) and a minority sensitive subgroup $S=1$ (e.g., race B). The ground-truth causal model is assumed to be as below:
$$
P(S=1)=0.1, P(S=0)=0.9,\epsilon_1,\epsilon_y,\epsilon_2 \sim \mathcal{N}(0,1),
$$
$$
X_1 \leftarrow S + U + \epsilon_1, Y \leftarrow X_1 + \epsilon_y, X_2 \leftarrow Y + \epsilon_2.
$$
As the subgroup $S=1$ is underrepresented, the fitted causal model may miss the causal relation $S\rightarrow X_1$ for $S=1$, i.e., the fitted causal model is biased (as the causal model $\mathcal{M}_2$ shown in Fig.~\ref{fig:example1}(c)). Then for \cfpa, $X_1$ and $X_2$ will be taken as input for prediction because they are considered to be non-descendants of $S$, but as $X_1$ and $X_2$ are actually biased because they are descendants of $S$, the predictor will also be biased consequently. Let us take the predictor $\hat{Y}=X_1$ for example. The predictor makes prediction $\hat{Y}_{S\leftarrow 0}=X_{1,S\leftarrow 0} = U+\epsilon_1$ and $\hat{Y}_{S\leftarrow 1}=X_{1,S\leftarrow 1} = U+\epsilon_1+1$ in when  $S\leftarrow 0$ and $S\leftarrow 1$, respectively, and this is obviously not counterfactually fair. For \cfpb, the fairness penalty on this biased causal model $\mathcal{M}_2$ is:
$$
\mathbb{E}(|f(X^{\mathcal{M}_2}_{S\leftarrow 1},1)-f(X^{\mathcal{M}_2}_{S\leftarrow 0},0)|) = |W_S|,
$$
while the fairness penalty based on the true causal model $\mathcal{M}$ is:
$$
\mathbb{E}(|f(X^{\mathcal{M}}_{S\leftarrow 1},1)-f(X^{\mathcal{M}}_{S\leftarrow 0},0)|) = |W_1+W_2+W_S|.
$$
Such difference may lead to inappropriate use of $X_1$ and $X_2$, and thus  bring biases to the predictor.

As a summary, existing counterfactual fairness machine learning methods heavily rely on given causal models, and would result in biases when the given causal models are incorrect.


\section{The Proposed Framework}
In this section, we introduce the proposed framework \mymodel, which targets at achieving counterfactual fairness without relying on explicit prior knowledge about the causal model. 
To achieve this goal, \mymodel~ learns counterfactually fair representations with counterfactual data augmentation, and then makes predictions based on the learned representations. 

\begin{figure*}[t]
\centering
  \begin{subfigure}[b]{0.23\textwidth}
        \centering
        \includegraphics[height=1.in]{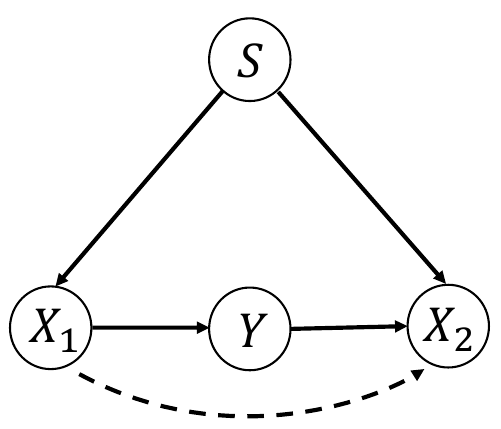}
        \caption{}
    \end{subfigure}
    \begin{subfigure}[b]{0.23\textwidth}
        \centering
        \includegraphics[height=1.in]{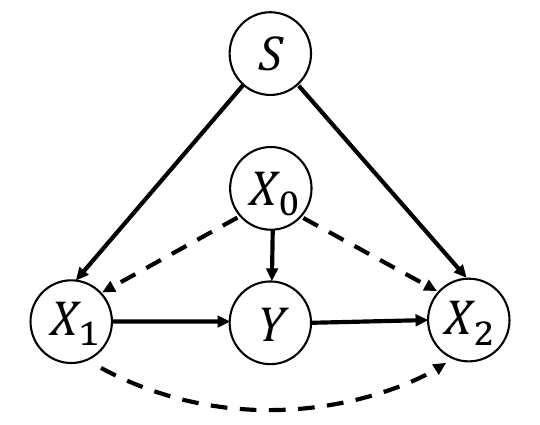}
        \caption{}
    \end{subfigure}
  \begin{subfigure}[b]{0.23\textwidth}
        \centering
        \includegraphics[height=1.in]{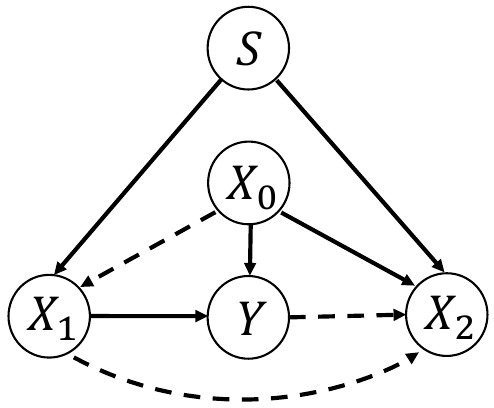}
        \caption{}
    \end{subfigure}
    \begin{subfigure}[b]{0.23\textwidth}
        \centering
        \includegraphics[height=1.in]{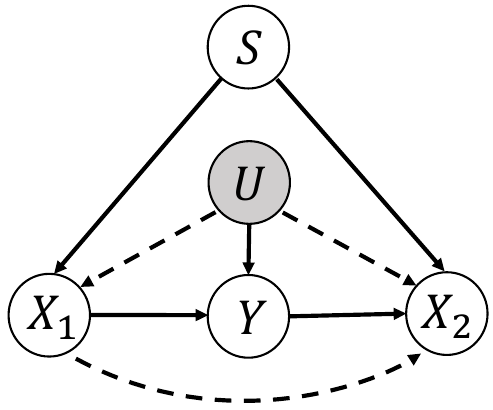}
        \caption{}
    \end{subfigure}
  \caption{Case studies of different kinds of variables in causal models. Each white (gray) node denotes an observed (unobserved) variable, each arrow denotes a causal relationship, and each dashed arrow denotes a possible causal relationship. $S,Y,U$ denotes the sensitive attribute, the prediction target, and the unobserved variable, respectively. $X_1$ is a causal variable of $Y$ and is a descendent of $S$, $X_0$ is a causal variable of $Y$ and is non-descendent of $S$, and $X_2$ is a variable with spurious correlations to $Y$.}
  \label{fig:casestudy}
\end{figure*}

\subsection{Assumptions and Examples}
Before technical details, we first present the key concepts and assumptions of \mymodel, and then use general examples of causal models (Fig.~\ref{fig:casestudy}) to describe the information needed in \mymodel. 

Previous works of counterfactual fairness \cite{kusner2017counterfactual} have discussed three levels of required prior knowledge about the causal model: 1) Level 1 only requires to know which observed features are non-descendants of the sensitive attribute, and only uses them for prediction; 
2) Level 2 postulates and infers the unobserved variables with partial prior knowledge of the causal model, and also uses them for prediction; 
3) Level 3 makes assumptions on the causal model (e.g., additive noise model \cite{hoyer2008nonlinear}), postulates the complete causal model, and 
then uses the inferred unobserved/observed non-descendants of the sensitive attribute for prediction. 
These three levels make increasingly stronger assumptions on the underlying causal model. But even the first level still requires to figure out which variables are non-descendants of the sensitive attribute. In this work, we aim to propose a principled way for counterfactually fair prediction without relying on the prior knowledge of the causal model. The main assumptions in our framework are listed as follows:
%

\begin{assumption} The sensitive attribute is not causally influenced by any other variables. This is a common assumption in most of existing fairness works \cite{kusner2017counterfactual,russell2017worlds,berk2018fairness}, as the commonly-used sensitive attributes such as race and gender usually do not have any causes.
\end{assumption}
\begin{assumption} If a variable $X_c$ directly affects $Y$ (i.e., an edge $X_c\rightarrow Y$ exists in the causal model), we assume $P(Y|X_c)$ is stable across different sensitive subgroups, but for the variables $X_s$ which do not causally affect $Y$, $P(Y|X_s)$ may be unstable in different sensitive subgroups. This assumption and its variants are widely used in invariant  learning \cite{arjovsky2019invariant,ahuja2020invariant}. 
\end{assumption}
%

As the ground truth causal model can be complicated, to investigate more general settings, we consider several different types of variables in the causal model, including descendant and non-descendant variables of $S$, causal and non-causal variables of $Y$, and observed and unobserved variables. Here we conduct several case studies on the causal model, and each corresponds to a causal graph shown in Fig.~\ref{fig:casestudy}. 
Suppose there is a ground truth causal model $\mathcal{M}$, we call the variables in $\mathcal{M}$ which causally affect the prediction target $Y$ (i.e., $Y$ is the descendant of such variables) as \textit{causal variables} of $Y$. 
In all the causal models in Fig.~\ref{fig:casestudy}, $X_1$ is a causal variable of $Y$, but it is also a descendant of $S$, thus it can not be directly used for counterfactually fair prediction. As shown in Fig.~\ref{fig:casestudy}(b) and (c), $X_0$ is also a causal variable of $Y$, and is non-descendant of $S$, thus $X_0$ is supposed to be used for fair prediction. 
$X_2$ is not a causal variable of $Y$, but it has statistically spurious correlations to $Y$. The reason may be that $X_2$ is $Y$'s descendant, as shown in Fig.~\ref{fig:casestudy}(b), or $X_2$ and $Y$ are affected by some common variables, as shown in Fig.~\ref{fig:casestudy}(c). As discussed in \cite{arjovsky2019invariant,chang2020invariant}, the spurious correlations between $X_2$ and $Y$ often vary across different sensitive subgroups and thus degrade the model prediction performance. Besides, if these non-causal variables are also descendants of sensitive attribute, incorporating them into prediction would also impede counterfactual fairness. Therefore, in our framework, we exclude these non-causal variables to further avoid potential biases. 
Above cases are all about observed variables, for those unobserved variables which are causative to $Y$, such as $U$ in Fig.~\ref{fig:casestudy}(d), we try to better capture these unobserved variables by utilizing the observed variables which have correlations with them.
%

Overall, in our framework, we \textit{learn representations to capture the causal variables which are not influenced by the sensitive attribute}.

\subsection{Overview of \mymodel~ Framework}
Existing counterfactual fairness works \cite{kusner2017counterfactual,russell2017worlds} involve counterfactual inference for predictor training, but it is often infeasible in real-world applications due to the lack of a correct causal model, especially when the data is noisy and high-dimensional \cite{belloni2017program}. 
Without enough knowledge about the causal model, inferring the unobserved variables and learning a fair predictor can be quite challenging. 
Here, we define the goal of our framework with respect to counterfactual fairness, and show an overview of the methodology. 



Based on the aforementioned preliminaries, we know that the key point of this problem is to capture the information which elicits a fair predictor, such as the causal variables that are non-descendants of $S$. In our framework, we use the observed features to learn a representation $Z=\Phi(X)$ which captures the fair information, and then build a predictor $\hat{Y}=g(Z)$ on top of it. 
In the implementation, we learn the representations $Z$ in the following ways: (1) To capture the causal variables of $Y$, we leverage the invariant risk minimization loss \cite{arjovsky2019invariant} to exclude those non-causal variables with unstable spurious correlations to $Y$. (2) To avoid taking the biases from the sensitive attribute into prediction, we develop a counterfactual data augmentation module, and encourage the learned representation to achieve the following goal: 
for any $s\ne s'$, and any $x$, 
$
    P(\Phi(x_{S\leftarrow s}))=P(\Phi(x_{S\leftarrow s'})).
$
Intuitively, it means that for each individual with observed features $x$ and sensitive attribute value $s$, the distributions of the representations learned from its original version and its counterfactuals should be the same.

\SetKwComment{Comment}{/* }{ */}
\begin{algorithm}[t]
\caption{The proposed \mymodel~ framework}\label{alg:framework}
\KwData{Instances of observable variables $\{X,S,Y\}$}
\KwResult{Counterfactually fair predictor $\hat{Y}=f(X,S)$}
/*~1. Counterfactual Data Augmentation~*/ \\
Train a VAE with encoder $\Psi(\cdot)$ and decoder $D(\cdot)$ with loss function in Eq.~(\ref{eq: loss_mmd}) (\mymodela) or Eq.~(\ref{eq:loss_adversarial}) (\mymodelb) \\
\For {each instance of random variables $\{X,S,Y\}$}{
Generate $K$ samples ${H}^1,...,{H}^K$ with $H = \Psi(X,Y)$ \\
\For {$s \in \mathcal{S}$}{
    $X^{CF}_{s}, Y^{CF}_{s} = \text{\textsc{Aggregate}}(D({H}^1,s),...,D({H}^K,s))$
}
} 
/*~2. Fair representation learning~*/ \\
Train a model $f=g\circ\Phi$ consisting of a representation learner $\Phi(\cdot)$ and a predictor $g(\cdot)$ \\
\For {each instance of random variables $\{X,S,Y\}$}{
    $Z=\Phi(X)$, $\hat{Y}=g(Z)$ \\
    \For {$s\in\mathcal{S}$}{
    $Z^{CF}_{s}=\Phi(X^{CF}_{s})$, $\hat{Y}^{CF}_{s}=g(Z^{CF}_{s})$
    }
    Back-propagation with loss function in Eq.~(\ref{eq:loss_f})
}
\end{algorithm}

Algorithm 1 shows an overview of our framework, including counterfactual data augmentation and fair representation learning. Detailed techniques will be introduced in the following subsections.

\subsection{Counterfactual Data Augmentation}
The lack of counterfactual data is the essential challenge to achieve counterfactual fairness. Thus, we pretrain a counterfactual data augmentation module to generate counterfactuals for each instance by manipulating its sensitive attribute. Then, the augmented counterfactuals together with original data are utilized to learn fair representations. The counterfactual data augmentation module is based on a variational auto-encoder (VAE) \cite{kingma2014auto} with an encoder-decoder structure. Specifically, the encoder in the VAE takes $\{X,Y\}$ as input, encodes them into a latent embedding space, and then the decoder reconstructs the original data $\{X,Y\}$ with the embeddings $H$ (notice that the embedding $H$ is different from the representation $Z$ introduced in the previous subsection. $H$ is the output of the bottleneck layer of the VAE in counterfactual data augmentation to generate counterfactuals) and sensitive attribute $S$. Note that $S$ is only used as an input of the decoder to enable counterfactual generation in later steps. 
The reconstruction loss $\mathcal{L}_r$ is:
\begin{equation}
    \mathcal{L}_r = \mathbb{E}_{q(H|X,Y)}[-\log(p(X,Y|H,S))] + \text{KL}[q(H|X,Y)\|p(H)],
\end{equation}
where $p(H)$ is a prior distribution, e.g., standard normal distribution $\mathcal{N}(0,I)$. $\text{KL}[\cdot\|\cdot]$ is the Kullback-Leibler (KL) divergence. 

To generate counterfactuals with the embeddings $H$ and a manipulated  sensitive attribute value later, we need to capture more ``fair" generative factors (i.e., those generative factors which are not causal influenced by $S$) in the embeddings, i.e., in encoder, we remove the causal influence of the sensitive attribute on the embedding $H$. Based on Assumption 1, if there is no dependency between the embeddings and sensitive attribute, then the embeddings encode no descendants of sensitive attributes. 
Now, we introduce two different implementations to remove the causal effect of $S$ on $H$ by minimizing the dependency between them. These implementations include the distribution matching based \mymodel~(\mymodela) and the adversarial learning based \mymodel~(\mymodelb). 

\noindent\textbf{Distribution matching based \mymodel.}
To remove the influence of the sensitive attribute, we  use the distribution matching technique \cite{shalit2017estimating,long2015learning} on the embeddings for different sensitive subgroups. We refer this implementation as \mymodela.
In particular, we minimize the Maximum Mean Discrepancy (MMD) \cite{long2015learning,shalit2017estimating} among the embedding distributions of different sensitive subgroups. 

The loss function of training the counterfactual data augmentation model  with distribution matching is as below:
\begin{equation}
    \min \mathcal{L}_r + \alpha \frac{1}{N_p}\sum\nolimits_{s\ne s'}MMD(P(H|s), P(H|s')),
\label{eq: loss_mmd}
\end{equation}
where $N_p=\frac{|\mathcal{S}|\times(|\mathcal{S}|-1)}{2}$ is the number of pairs of different sensitive attribute values, and $|\mathcal{S}|$ is the number of different sensitive attribute values.  
The second term is the distribution matching penalty, which aims to achieve $P(H|S=s)=P(H|S=s')$ for all pairs of different sensitive subgroups $(s, s')$. Here $\alpha\ge0$ is a hyperparameter which controls the importance of the distribution balancing term. 

\noindent\textbf{Adversarial Learning based \mymodel.}
We also propose an adversarial learning based implementation, referred as~\mymodelb. In this implementation, we train a discriminator $h(\cdot)$ which uses the embeddings to distinguish instances that bear different values of the sensitive attribute. The objective function is  as below:
\begin{equation}
    \min_{\Psi(\cdot)}\max_{h(\cdot)}  \mathcal{L}_{r} + \alpha'\frac{1}{|\mathcal{S}|} \sum\nolimits_{s\in\mathcal{S}}\mathbb{E}_{X^s,S^s}[\log P(h(H)=s)],
\label{eq:loss_adversarial}
\end{equation}
where $\Psi(\cdot)$ is the encoder. The first term is the aforementioned reconstruction loss. The second term calculates the probability that the discriminator makes correct predictions for each instance's sensitive attribute. Therefore, the sensitive attribute predictor $h(\cdot)$ is playing an adversarial game with the encoder $\Psi(\cdot)$. In this way, the embeddings are encouraged to exclude the information related to the sensitive attribute. Here $\alpha'\ge0$ is a hyperparameter to control the weight of the sensitive attribute discriminator. The minimax problem is optimized with an alternating gradient descent process.

\subsection{Fair Representation Learning}
\subsubsection{Counterfactually Fair Representations} 
With the counterfactual data augmentation module, we generate counterfactuals by feeding the embeddings $H$ and a sensitive attribute value $s'$ different from the original one $s$ into the decoder ${D}(\cdot)$, and taking the output $(X^{CF}_{s'},Y^{CF}_{s'})=D(H,s')$ as the counterfactuals corresponding to  $S\leftarrow s'$. For each instance and each sensitive attribute value, we generate $K$ samples of embeddings $({H}^{1},...,{H}^{K})$, and aggregate the corresponding counterfactuals by an operation $\textsc{Aggregate}(\cdot)$ (e.g., mean). For notation simplicity, we still denote the aggregated counterfactual data as $(X^{CF}_{s'},Y^{CF}_{s'})=\textsc{Aggregate}(D({H}^1,s'),...,D({H}^K,s'))$. Based on these counterfactuals, we train a representation learner $\Phi(\cdot)$ which maps instance features $X$ into representations: $Z=\Phi(X)$, and we use a predictor $g(\cdot)$ to make predictions based on $Z$. 

To learn counterfactually fair representations $Z$, we add a counterfactual fairness constraint to mitigate the discrepancy between the representations learned from original data and its corresponding counterfactuals. The constraint is formulated as:
\begin{equation}
    \mathcal{L}_c =\! \frac{1}{|\mathcal{S}|-1}\sum_{s'\ne s} d(Z, Z^{CF}_{s'})=\!\frac{1}{|\mathcal{S}|-1}\sum_{s'\ne s} d(\Phi(X), \Phi(X^{CF}_{s'})),
\end{equation}
where $X^{CF}_{s'}$ is the counterfactual generated in counterfactual data augmentation corresponding to $S\leftarrow s'$, and $d(\cdot,\cdot)$ is a distance metric such as cosine distance to measure the discrepancy between two representations.

\subsubsection{Invariant Representations} 
As aforementioned, the non-causal variables which have spurious correlations to the target $Y$ are likely to degrade the model prediction performance, and may also incorporate potential biases from sensitive attributes to prediction.
It has been shown in~\cite{arjovsky2019invariant} that the relationships from these variables to $Y$ often vary across different domains, e.g., different sensitive subgroups.
%
Therefore, to exclude the influence of such non-causal variables on the learned representations and capture the causal variables of $Y$, we leverage the invariant risk minimization (IRM) loss~\cite{arjovsky2019invariant} for the sensitive subgroup $s$ as below:
\begin{equation}
\mathcal{L}_{IRM}^s =R^s(g\circ\Phi) + \lambda\left\|\bigtriangledown_{w|w=1.0}R^s(w\cdot (g\circ\Phi))\right\|_{2}^2,
\label{eq:irm1}
\end{equation}
where $\mathcal{L}^s_{IRM}$ is the IRM loss in the sensitive subgroup $s$, the first term $R^s(g\circ\Phi)=\mathbb{E}[\mathcal{L}(g(\Phi(X^s,S^s)),Y^s)]$ is the prediction loss under sensitive subgroup $s$, 
and $w$ is a scalar and is fixed as $w=1.0$. 
According to \cite{arjovsky2019invariant}, the gradient of $R^s(w\cdot (g\circ\Phi))$ w.r.t. $w$ can reflect the ``invariance" of the learned representations. 
Therefore, in the above formulation, the second term measures the invariance of the relationship between the representations and the target across different sensitive groups. 
Here, $\lambda$ is a hyperparameter for the trade-off between the prediction performance and the level of invariance. The IRM loss aims to ensure that the predictor can be optimal in all the different sensitive subgroups, thus the unstable spurious correlations varying across sensitive subgroups can be excluded.


To put it all together, the overall loss function for fair representation learning is as follows:
\begin{equation}
    \mathcal{L} = \frac{1}{|\mathcal{S}|}\sum\nolimits_{s\in\mathcal{S}}\mathcal{L}^s_{IRM} + \beta\mathcal{L}_c,
    \label{eq:loss_f}
\end{equation}
where $\beta$ is the weight of the counterfactual fairness constraint.
More implementation details can be found in Appendix A.

\section{Experimental Evaluations}
In this section, we conduct extensive experiments to evaluate the proposed framework \mymodel~ on two real-world datasets and one synthetic dataset. Before showing the detailed results, we first present the details of used datasets and the experimental settings. 



\subsection{Datasets}



\begin{figure}[t]
\centering
  \begin{subfigure}[b]{0.232\textwidth}
        \centering
        \includegraphics[height=1.in]{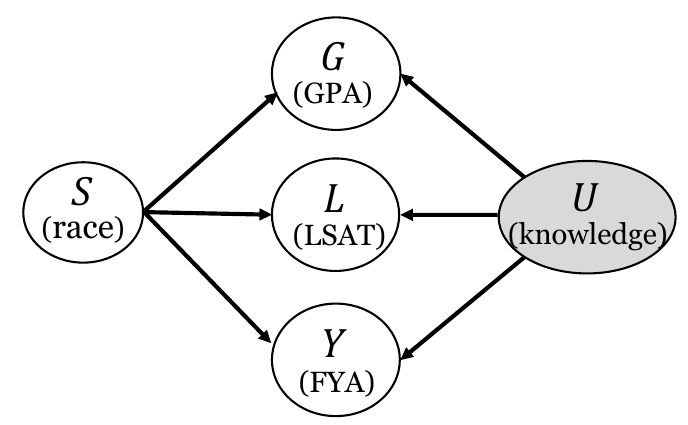}
        \caption{Law school}
    \end{subfigure}
  \begin{subfigure}[b]{0.24\textwidth}
        \centering
        \includegraphics[height=1.in]{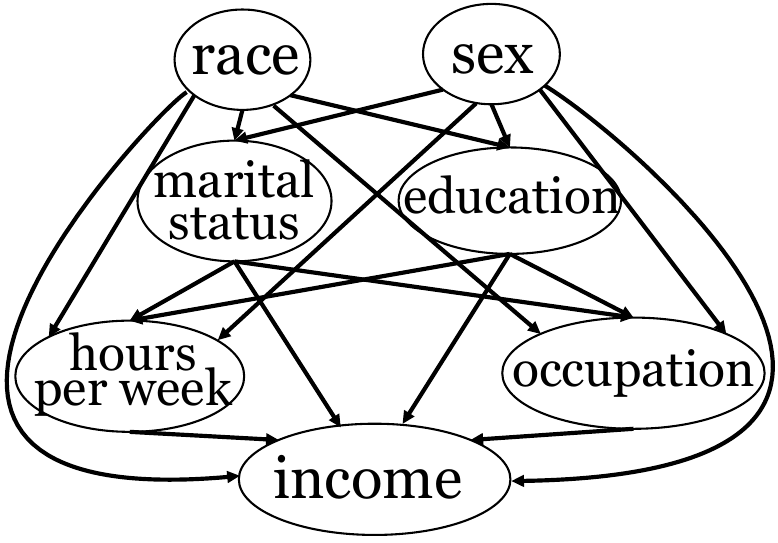}
        \caption{Adult}
    \end{subfigure}
  \caption{The ground truth causal models of two real-world datasets Law School and Adult.}
  \label{fig:causalmodel_real}
\end{figure}

\noindent \textbf{Law School.} This dataset 
contains academic information of students in $163$ law schools. 
Our goal is to predict each student's first year average grade (FYA), and this is a regression task.
We take \textit{race} as their sensitive attribute, and take grade-point average (GPA) and entrance exam scores (LSAT) as two observed features.  Here, we select persons in races of white, black, and asian. The dataset contains $20,412$ instances. 
We use the level-2 causal model in \cite{kusner2017counterfactual} as the true causal model with causal graph shown in Fig.~\ref{fig:causalmodel_real}(a).

\noindent \textbf{Adult.} UCI Adult income dataset\footnote{https://archive.ics.uci.edu/ml/datasets/adult} contains census data for different adults and the target here is to predict whether their income exceeds 50K/yr. We take \textit{race} as the sensitive attribute $S$, and their \textit{income} as the prediction label $Y$. This is a binary classification task. We select persons in the races of white, black, and Asian-Pac-Islander. In addition to the sensitive attribute of \textit{race}, we use other $5$ attributes for prediction. The dataset contains $31,979$ instances. Here, we follow \cite{wu2019counterfactual} and consider the causal model used by them as the ground truth. The causal graph is shown in Fig.~\ref{fig:causalmodel_real}(b).

\begin{figure}[t]
\centering
\begin{subfigure}[b]{0.15\textwidth}
        \centering
        \includegraphics[height=1.1in]{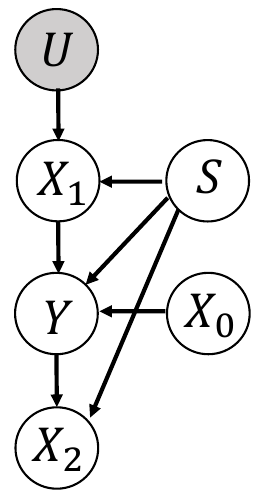}
        \caption{$\mathcal{M}$}
    \end{subfigure}
    \begin{subfigure}[b]{0.15\textwidth}
        \centering
        \includegraphics[height=1.1in]{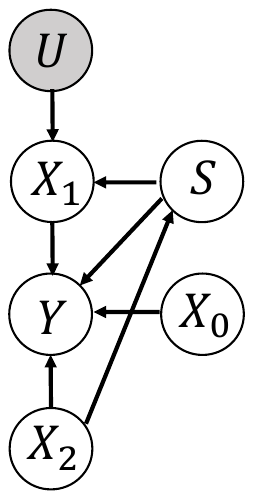}
        \caption{$\mathcal{M}_1$}
    \end{subfigure}
    \begin{subfigure}[b]{0.15\textwidth}
        \centering
        \includegraphics[height=1.1in]{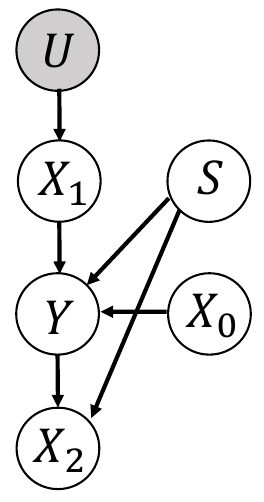}
        \caption{$\mathcal{M}_2$}
    \end{subfigure}
  \caption{The true causal model ($\mathcal{M}$) and two incorrect causal models ($\mathcal{M}_1$ and $\mathcal{M}_2$) of the synthetic dataset.}
  \label{fig:causalmodel_sythetic}
\end{figure}

\noindent \textbf{Synthetic Dataset.} Here, we use a ground truth causal model to generate the synthetic data. The true causal graph is shown in Fig.~\ref{fig:causalmodel_sythetic}(a), containing a sensitive attribute $S$ with four different categorical values $\{0,1,2,3\}$, an unobserved variable $U$, a causal variable $X_0$ which is non-descendant of $S$, a causal variable $X_1$ which is descendant of $S$, and a variable $X_2$ which is the descendant of $Y$. The structural equations are as follows:
$$
S \sim \text{Catgorical}(\pi), U\sim \mathcal{N}(0,\sigma_U^2),X_0 = \mathcal{N}(0,\sigma_0^2),
$$
$$
X_1 = W_S S + U + \mathcal{N}(0,\sigma_{S,1}^2), Y = X_1 + X_0 + \mathcal{N}(0,\sigma_{S,Y}^2),
$$
\begin{equation}
    X_2 = Y + \mathcal{N}(0,\sigma_{S,2}^2), 
    \label{eq:synthetic}
\end{equation}
where $\pi=\{0.5,0.4,0.05,0.05\}$, $\sigma_U=\sigma_0=1$, $\sigma_{S,*}$ and $W_S$ are set as $\{0.5, 1.0, 1.5, 2.0\}$ and $\{0.1,0.2,1.0,2.0\}$ respectively for four values of sensitive attribute. In this dataset, the spurious correlation $X_2 \rightarrow Y$ and the imbalanced distribution of sensitive subgroups may lead to incorrect causal models, as shown in \cite{nauta2019causal}. We will further investigate the impact of these two situations in Section 4.4.


\vspace{2mm}
\subsection{Experimental Settings}

\noindent \textbf{Baselines.} To investigate the effectiveness of our framework in learning counterfactually fair predictors from observational data, we compare the proposed framework with multiple state-of-the-art methods. First, we briefly introduce all the compared baseline methods and their settings: 
\begin{itemize}
    \item \textbf{Constant Predictor:} A predictor which has constant output for any input. We obtain this constant predictor by finding a constant which can minimize the mean squared error (MSE) loss on the training data. 
\item  \textbf{Full Predictor:} Full predictor takes \textit{all} the observed attributes (except the attribute used as label) as input for prediction. 
\item \textbf{Unaware Predictor:} Unaware predictor is based on the notion of fairness through unawareness~\cite{grgic2016case}. It takes all features except the sensitive attribute as input to predict the label. 
\item  \textbf{Counterfactual Fairness Predictor:} We use two different counterfactual fairness predictors here, including \textbf{\cfpa}~  \cite{kusner2017counterfactual} and  \textbf{\cfpb}~ \cite{russell2017worlds}. These methods require a given causal model. 
\end{itemize}

For baselines full/unaware/counterfactual fair predictors, we use linear regression for regression and logistic regression for classification. 
More details of baselines can be found in Appendix B.

\begin{table*}[t]
\centering
 \caption{Results comparison of different predictors on two real-world datasets. Our method \mymodel~ can achieve the best performance in counterfactual fairness with competitive prediction performance.}
  \def\arraystretch{1.2}%
 \label{tab:prediction}
  \begin{tabular}{l||cccc||cccc}\hline
    \multirow{2}{*}{Method} & \multicolumn{4}{|c||}{Law school} &  \multicolumn{3}{c}{Adult} \\ \cline{2-8} 
     & RMSE ($\downarrow$)  & MAE ($\downarrow$)  & MMD ($\downarrow$) & Wass($\downarrow$) &  Accuracy ($\uparrow$)  & MMD ($\downarrow$) & Wass ($\downarrow$)  \\
    \hline
    Constant & $0.952\pm 0.003$ & $0.772\pm 0.002$ & $\bm{0.000\pm 0.000}$ & $\bm{0.000\pm 0.000}$ & $0.745 \pm 0.001$  & $\bm{0.000\pm 0.000}$ &  $\bm{0.000\pm 0.000}$ \\
    Full  & \bm{$0.896 \pm 0.004$}   & \bm{$0.723 \pm 0.003$} & $259.744 \pm 5.213$ & $65.656 \pm 1.326$ & \bm{$0.815 \pm 0.002$} & $50.513\pm 3.283$ & $5.217 \pm 0.582$\\
    Unaware & \underline{$0.909 \pm 0.002$} & $0.734 \pm 0.004$ & $39.144 \pm 3.248$ & $10.093 \pm 1.254$ & $0.809 \pm 0.003$ & $16.832\pm 2.377$ & $1.983 \pm 0.462$\\
    \hline
    \cfpa~(true) & $0.932 \pm 0.003$ & $0.738 \pm 0.002$ & $4.307 \pm 0.003$ & \underline{$0.019 \pm 0.001$} & $0.745 \pm 0.002$ & $3.582 \pm 0.007$ & $0.025\pm 0.002$\\
    \cfpb~(true)& $0.929 \pm 0.004$ & $0.735 \pm 0.003$ & $4.325 \pm 0.002$ &$0.020 \pm 0.012$ & $0.748 \pm 0.003$&  $3.623 \pm 0.004$& $0.029\pm 0.004$ \\
    \hline
    \mymodela~(ours) & \underline{$0.909 \pm 0.002$} & \underline{$0.733 \pm 0.003$} & $\underline{4.297 \pm 0.002}$ & \underline{$0.019 \pm 0.001$} & $0.778 \pm 0.002$& \underline{$3.552 \pm 0.021$} & \underline{$0.023\pm 0.002$}\\
    \mymodelb~(ours) & $0.910 \pm 0.002$& $0.734 \pm 0.002$& \underline{$4.289 \pm 0.002$}& \underline{$0.018 \pm 0.001$} & $0.780 \pm 0.003$& \underline{$3.547\pm 0.007$}& \underline{$0.023 \pm 0.002$}\\
    \hline
\end{tabular}
\end{table*}

\begin{table}[t]
\small
\centering
 \caption{Study on synthetic data about the adverse effects of incorrect causal model $\mathcal{M}_1$.} 
 \label{tab:spurious}
 \def\arraystretch{1.2}%
  \begin{tabular}{l||cc||cc}
\hline
Method     & RMSE  & MAE  & MMD & Wass \\
    \hline
\cfpa~(true)    & $1.34 \pm 0.01$ & $0.88 \pm 0.01$ & $8.42 \pm 0.70$  & $3.07 \pm 0.01$ \\
\cfpa~ (false) & $\underline{1.30 \pm 0.01}$ & $\underline{0.83\pm 0.02}$  & $10.11\pm0.52$   & $3.79\pm 0.03$  \\
\cfpb~(true)    & $1.32 \pm 0.01$ & $0.87 \pm 0.01$ & $8.48 \pm 0.83$  & $3.32 \pm 0.02$ \\
\cfpb~(false) & $\bm{1.29\pm0.01}$   & $\bm{0.81\pm0.01}$   & $10.94\pm 0.61$  & $3.84\pm0.02$   \\
\hline
\mymodel-M   & $1.32 \pm 0.01$ & $0.87 \pm 0.02$ & \underline{$7.52\pm 0.08$}   & \underline{$2.63\pm 0.02$}  \\
\mymodel-A   & $1.31 \pm 0.01$ & $0.85 \pm 0.03$ & \bm{$7.49\pm0.05$}    & \bm{$2.58\pm0.01$}  \\
    \hline
\end{tabular}
\end{table}

\begin{table}[t]
\centering
\small
 \caption{Study on synthetic data regarding the adverse effects of incorrect causal model $\mathcal{M}_2$.} 
 \label{tab:imbalance}
 \def\arraystretch{1.2}%
  \begin{tabular}{l||cc||cc}
    \hline
    \multirow{2}{*}{Method} & \multicolumn{2}{c||}{$S\leftarrow 0$ and $S\leftarrow 1$} &  \multicolumn{2}{c}{$S\leftarrow 0$ and $S\leftarrow 2$} \\ \cline{2-5}
     & MMD  & WASS  & MMD & Wass  \\
    \hline
    \cfpa~(true) & $\underline{6.05 \pm 0.02}$ &\underline{$1.10\pm 0.02$}  & $7.97 \pm 0.03$ & $2.55\pm 0.02$   \\
    \cfpa~(false) &$6.63 \pm 0.09$  & $1.24 \pm 0.04$ & $9.33 \pm 1.00$  & $3.62 \pm 0.01$ \\
    \cfpb~(true) &$6.34 \pm 0.07$ & $1.13\pm0.03$   & $8.31\pm 0.98$  & $2.84\pm 0.03$ \\
    \cfpb~(false) &$6.83 \pm 0.08$  & $1.35 \pm 0.05$ & $9.92 \pm 1.01$  & $3.98 \pm 0.02$ \\
    \hline
    \mymodela & \underline{$6.12 \pm 0.04$}  & \underline{$1.13 \pm 0.02$} & \underline{$7.94 \pm 0.06$}  & \underline{$2.52 \pm 0.01$}  \\
    \mymodelb & \bm{$6.05 \pm 0.03$}  & \bm{$1.11 \pm 0.03$} & \bm{$7.42 \pm 0.04$}  & \bm{$2.49\pm 0.01$}\\
    \hline
\end{tabular}
\end{table}

\noindent \textbf{Evaluation Metrics.}
Generally, the evaluation metrics consider two different aspects: prediction performance and counterfactual fairness. To measure the model prediction performance, we employ the commonly used metrics -- Root Mean Square Error (RMSE) and mean absolute error (MAE) for regression tasks and accuracy for classification tasks. 
To evaluate different methods with respect to counterfactual fairness, we compare the distribution divergence of the predictions made on different counterfactuals generated by the ground truth causal model. 
If a predictor is counterfactually fair, the distributions of the predictions under different ground-truth counterfactuals are expected to be the same. Here, we use two distribution distance metrics (including Wasserstein-1 distance (Wass) \cite{ruschendorf1985wasserstein} and Maximum Mean Discrepancy (MMD) \cite{long2015learning,shalit2017estimating}) to measure the distribution divergence. 
We compute the divergence of prediction distributions in every pair of counterfactuals ($S\leftarrow s$ and $S\leftarrow s'$ for any $s\ne s'$), then take the average value as the final result. The smaller the average values of MMD and Wass are, the better a predictor performs in counterfactual fairness. For the synthetic data, the ground truth causal model is known, while for the real-world datasets, we adopt the widely accepted causal models as mentioned in Section 4.1. 


\noindent \textbf{Hyperparameter Settings.} For all these three datasets, we split the training/validation/test set as $60\%/20\%/20\%$. All the presented results are on the test data. We set the number of training epochs as $500$, the representation dimension as $10$, $\alpha=2.0$, $\alpha'=1.0$, $K=20$, $\beta=5.0$, and $\lambda=1.0$. 



\subsection{Experimental Results on Real-world Data}
To assess the superiority of the proposed framework \mymodel, we compare its two implementations \mymodela~ and \mymodelb~ against other predictors on two real-world datasets Law School and Adult. We show the ground truth causal models of these two datasets in Fig.~\ref{fig:causalmodel_real} although our proposed framework and its variants do not rely on the causal model. Table \ref{tab:prediction} presents the performance of different methods regarding prediction and counterfactual fairness. The best results are shown in \textbf{bold}, and the runner-up results are \underline{underlined}.
Generally speaking, existing methods which are not designed for counterfactual fairness have higher MMD and Wass, although they can use the biased features to achieve better prediction performance. 
We make the following observations from Table~\ref{tab:prediction}:
\begin{itemize}
\item Among all the compared methods, the constant predictor has the worst performance in prediction as it lacks capability to distinguish different instances. However, it always satisfies counterfactual fairness because it has constant output. 
\item The full predictor performs well in prediction, as it utilizes all the features (both sensitive and non-sensitive). But the use of sensitive attribute also brings biases to the prediction, as demonstrated by its high values on fairness metrics. 
\item The unaware predictor removes certain biases by ignoring the sensitive attribute, but it cannot exclude the implicit biases caused by inappropriate usage of the descendants of the sensitive attribute. 
\item Both \cfpa~and \cfpb~ infer the latent variables based on the given causal model, so they perform well if the given causal model is correct. 
\item Our proposed \mymodel~consistently outperform other baselines (except the constant predictor) under different fairness metrics, and also have better prediction performance than many other fairness-aware baselines (including \cfpa~ and \cfpb). It implies that \mymodel~ can achieve a good balance between prediction performance and counterfactual fairness.
\item The variants \mymodela~and~\mymodelb~ generally have similar performance, but \mymodelb~ is slightly better in fairness, it may benefit from the effectiveness of its adversarial learning mechanism in removing the sensitive information.
\end{itemize}

\vspace{0.05in}
\subsection{Experimental Results on Synthetic Data}
The above experiments on real-world datasets have demonstrated the superiority of \mymodel. Here, we perform further studies on the synthetic dataset to show the impact of incorrect causal models. 

\noindent \textbf{Incorrect causal model $\mathbf{M_1}$.} %
In this experiment, we use the synthetic data to showcase the impact of an incorrect causal model as the example shown in Fig. \ref{fig:causalmodel_sythetic}(b). 
The true causal model of the synthetic data is shown in Fig. \ref{fig:causalmodel_sythetic}(a). 
Here, causal relations regarding $X_2$ in $\mathcal{M}_1$ are reversed. 
As all the baselines (except \cfpa~and \cfpb) do not rely on the causal model for prediction, so their results are not influenced by the correctness of the causal model. Here, we investigate the influence of the incorrect causal model on \cfpa~and \cfpb~ and compare their performance with our proposed framework. 
%
From the results shown in Table~\ref{tab:spurious}, 
we find the fairness of \cfpa~ and \cfpb~ are obviously affected by the incorrect causal model.
Although \cfpa~ and \cfpb~ with incorrect causal model have slightly better performance in prediction, that is because based on the incorrect causal model, they may take $X_2$ into prediction, which however, brings biases for prediction. 
Our proposed framework does not assume the existence of any given causal model for prediction. The counterfactual data augmentation enables us to eliminate the influence of sensitive attributes to the prediction. Furthermore, the learned invariant representations in \mymodel~ exclude the adverse impacts of non-causal variables with spurious correlations and leverage the causal variables to learn representations, thus $X_2$ is encouraged to be excluded from prediction.


\noindent \textbf{Incorrect causal model $\mathbf{M_2}$.} Now, we use the  synthetic data to showcase the impact of another incorrect causal model as shown in Fig.~\ref{fig:causalmodel_sythetic}(c). 
%
As described in Section 4.1, we set the parameter $W_S$ in Eq.~(\ref{eq:synthetic}), which determines the relation $S\rightarrow X_1$, to be small on the majority sensitive subgroups ($S=0,1$) but relatively large on the minority sensitive subgroups ($S=2,3$). 
Here, the incorrect causal model misses the causal relation $S\rightarrow X_1$ (as shown in Fig.~\ref{fig:causalmodel_sythetic}(c)). 
%
We compare the prediction differences between pairs of different counterfactuals generated by the true causal model shown in Fig.~\ref{fig:causalmodel_sythetic}(a). 
The results are shown in Table \ref{tab:imbalance}, 
where we select two pairs of counterfactuals: $(S\leftarrow 0$ and $S\leftarrow 1)$ and $(S\leftarrow 0$ and $S\leftarrow 2)$. 
As $W_S$ is small when $S=0$ and $S=1$, the biased causal model would not bring too much bias from the sensitive attribute to the prediction in the two counterfactuals $(S\leftarrow 0$ and $S\leftarrow 1)$, so the discrepancy between this pair is relatively lower than the other pair. But for the counterfactuals of $S\leftarrow 2$ (and also $S\leftarrow3$), \cfpa~ and \cfpb~ suffer more from the biased causal model. As observed in Table \ref{tab:imbalance},
when \cfpa~ and \cfpb~ are under the biased causal model, the prediction discrepancy between the pair of counterfactuals $(S\leftarrow 0$ and $S\leftarrow 2)$ becomes larger than the case when \cfpa~ and \cfpb~ are under the true causal model. 
Similar observations can also be found in the pair $(S\leftarrow 2$ and $S\leftarrow 3)$, as shown in Appendix C. 
Our framework outperforms the baselines due to the following key factors: the fair generative factors captured in  counterfactual data augmentation remove the influence of the observed sensitive attribute to the generated counterfactuals. Therefore, the counterfactual fairness constraint 
mitigates the influence of sensitive attribute on the learned representations, and  makes our framework suffer less from  imbalanced sensitive subgroups.

\begin{figure}[t]
\centering
  \begin{subfigure}[b]{0.24\textwidth}
        \centering
        \includegraphics[height=1.1in]{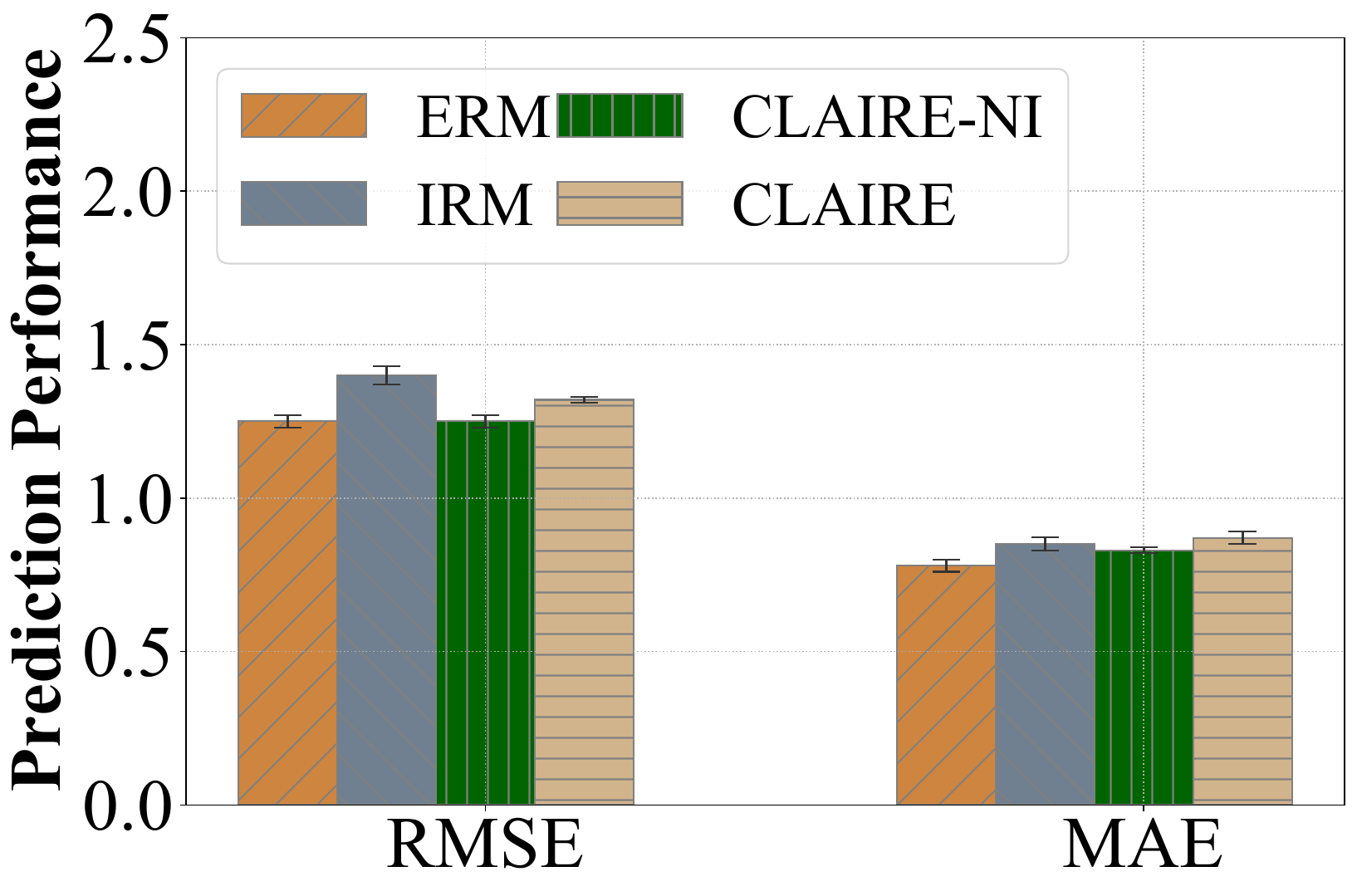}
        \caption{Prediction}
    \end{subfigure}
  \begin{subfigure}[b]{0.23\textwidth}
        \centering
        \includegraphics[height=1.1in]{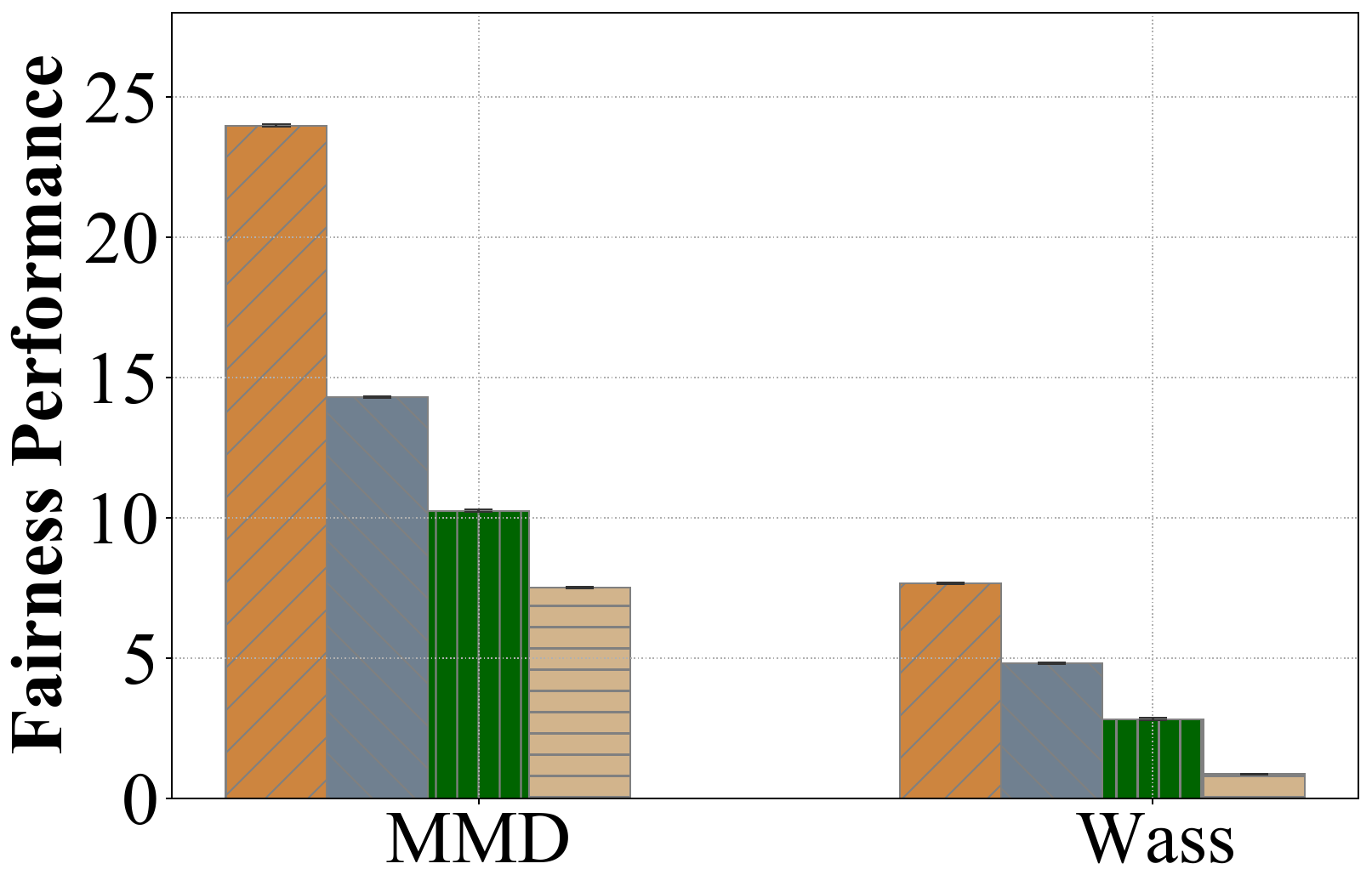}
        \caption{Fairness}
    \end{subfigure}
  \caption{Ablation Study on Synthetic Dataset.}
  \label{fig:ablation}
\end{figure}

\begin{figure}[t]
\centering
  \begin{subfigure}[b]{0.23\textwidth}
        \centering
        \includegraphics[height=1.1in]{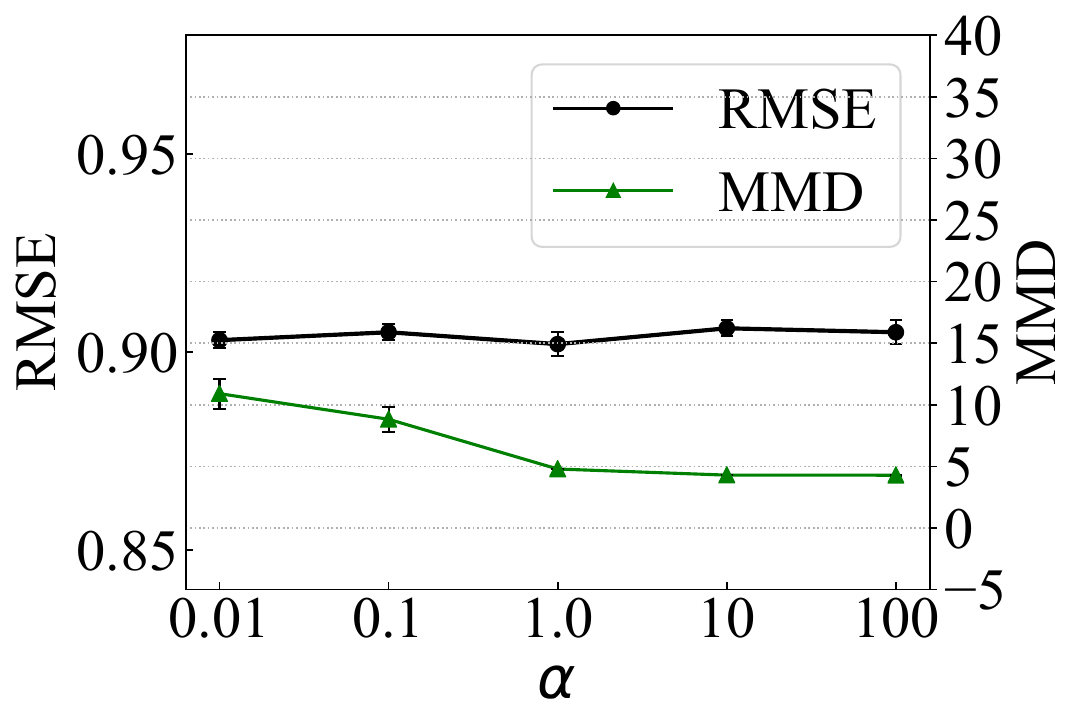}
        \caption{Vary $\alpha$}
    \end{subfigure}
  \begin{subfigure}[b]{0.23\textwidth}
        \centering
        \includegraphics[height=1.1in]{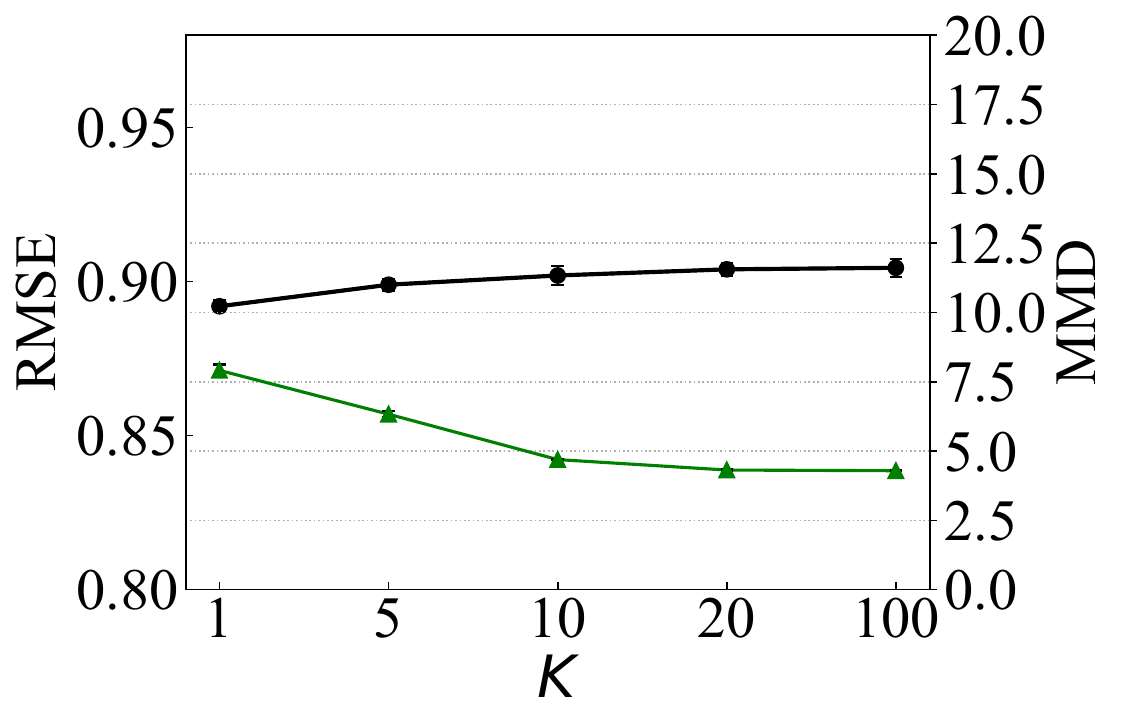}
        \caption{Vary $K$}
    \end{subfigure}
     \begin{subfigure}[b]{0.23\textwidth}
        \centering
        \includegraphics[height=1.1in]{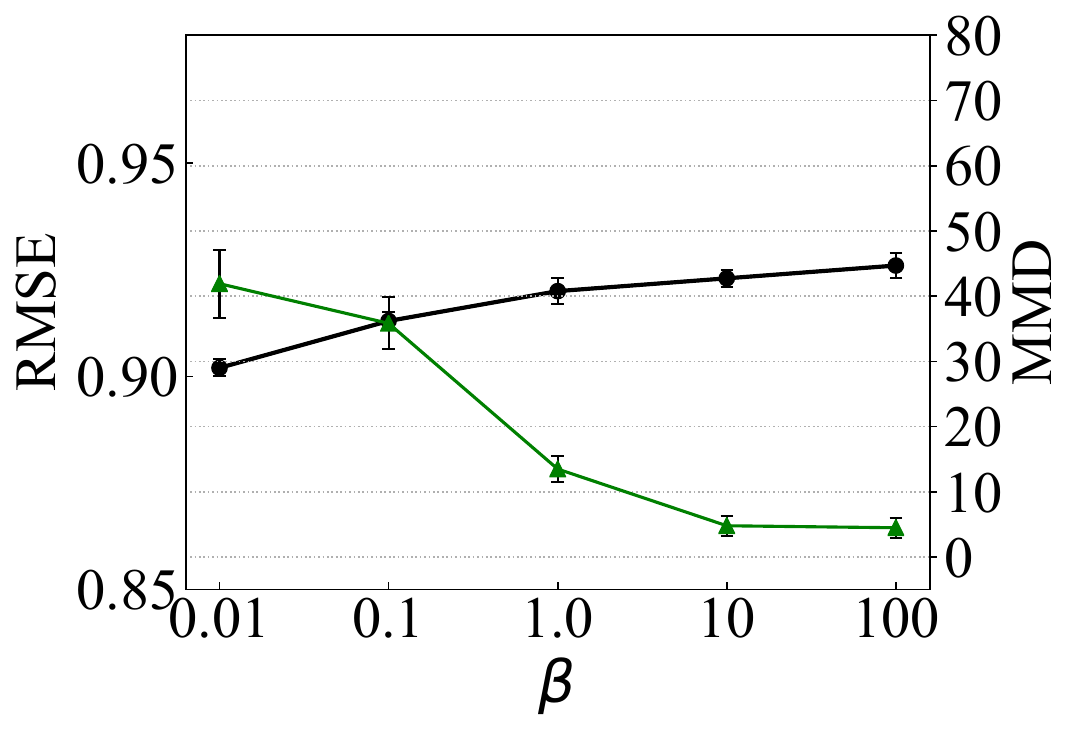}
        \caption{Vary $\beta$}
    \end{subfigure}
  \begin{subfigure}[b]{0.23\textwidth}
        \centering
        \includegraphics[height=1.1in]{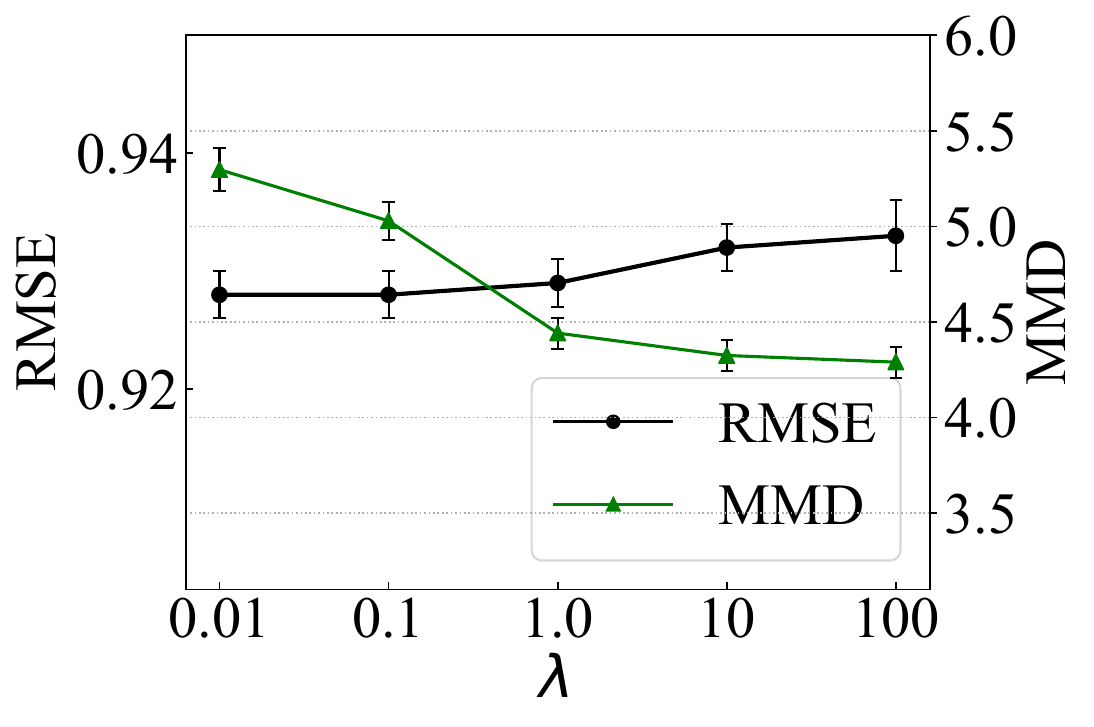}
        \caption{Vary $\lambda$}
    \end{subfigure}
   \vspace{-3mm}
  \caption{Performance of \mymodel~with different settings of hyperparameters.}
  \label{fig:param}
\end{figure}

\subsection{Ablation Study}
To evaluate the effectiveness of each component in our method, we provide ablation study with the following variants:
1) \textbf{Empirical Risk Minimization (ERM):} ERM can be considered as a variant of our proposed framework \mymodel. Here, we only use the empirical risk minimization loss (the first term of Eq.~(\ref{eq:irm1})) in prediction without the counterfactual fairness constraint and invariant penalty by setting $\beta=0$ and $\lambda=0$. 
2) \textbf{Invariant Risk Minimization (IRM) \cite{arjovsky2019invariant}:} Here, we remove the counterfactual fairness constraint in our framework by setting $\beta=0$. 
3) \textbf{\mymodel-NI:} As the third variant of our proposed framework, we remove the invariant penalty by setting $\lambda=0$ in \mymodel. From the results shown in Fig.~\ref{fig:ablation}, the counterfactual data augmentation and invariant penalty both contribute to the overall fairness performance.

\subsection{Parameter Study}
We set the hyperparameter $\alpha\in\{0.01, 0.1, 1.0, 10, 100\}$, the sampling number $K\in\{1, 5, 10, 20, 100\}$, $\beta\in\{0.01, 0.1, 1.0, 10, 100\}$, $\lambda\in\{0.01, 0.1, 1.0, 10, 100\}$, and compare the performance of our proposed framework in Fig.~\ref{fig:param}. Here we only show the results of \mymodela~ on the law school dataset, as similar patterns can be observed in \mymodelb~ and other datasets. As observed in  Fig.~\ref{fig:param}(a), $\alpha$ controls the ``fairness" of the embedding in counterfactual data augmentation. Larger values of $\alpha$ can improve the counterfactual fairness of the framework, and have no obvious impact on the prediction performance. With larger $K$ in Fig.~\ref{fig:param}(b), the performance of counterfactual fairness also improves because more samples are generated in counterfactual data augmentation.
$\beta$ controls the importance of counterfactual fairness constraint, $\lambda$ controls the invariance penalty of the representations. 
%
As shown in Fig. \ref{fig:param}(c), with the increase of $\beta$, the framework focuses more on removing the biases from the sensitive attribute, which may sacrifice some information to predict the target, and thus results in higher RMSE, but can achieve better fairness. As shown in Fig. \ref{fig:param}(d), with the increase of $\lambda$, the framework may exclude more variables with unstable relationships to the target across different sensitive subgroups, it may thus lose some information specific to each sensitive subgroup, but can also contribute to better fairness.
From the observations, the framework achieves a good trade-off on the prediction performance and counterfactual fairness with proper parameter settings. 
\section{Related Work}

\noindent \textbf{Counterfactual Fairness.} 
Recently, aside from traditional statistical fairness notions \cite{zemel2013learning,hardt2016equality,zafar2017fairness,dieterich2016compas,chouldechova2017fair,chouldechova2018frontiers,barocas2017fairness}, 
causal-based fairness notions \cite{makhlouf2020survey,kusner2017counterfactual,russell2017worlds} have attracted a surge of attentions because of its strong capability of modeling how the discrimination is exhibited.
Among them, the notion of counterfactual fairness \cite{kusner2017counterfactual} assesses fairness at the individual level. 
Most of the existing counterfactual fairness studies \cite{xu2019achieving,kusner2017counterfactual,grari2020adversarial} are based on a given ground-truth causal model or rely on causal discovery methods \cite{spirtes2000causation,pearl2009causality,kalisch2007estimating}. Multi-world fairness \cite{russell2017worlds} considers the situation when the ground-truth causal model cannot be decided, but it still requires a candidate set containing causal models which may be true, and proposes an optimization based method to achieve counterfactual fairness with the average of the causal models in the candidate set. 
Many methods based on traditional causal discovery are limited in certain scenarios, such as low-dimensional and linear settings. 
Recent studies \cite{kim2021counterfactual,grari2022adversarial,zuo2022counterfactual} provide more discussion about counterfactual fairness under different assumptions and scenarios. 
But in conclusion, most of the above methods require much explicit prior knowledge of the causal model to remove the influence of the sensitive attribute on the prediction, and lack discussion of the impact of incorrect causal models. 

\noindent \textbf{Invariant Risk Minimization.} Invariant risk minimization (IRM) \cite{arjovsky2019invariant} and its variants \cite{guo2021out,ahuja2020invariant,krueger2020out,jin2020domain,mahajan2020domain,chang2020invariant} are originally proposed for out-of-distribution (OOD) generalization \cite{krueger2020out,sagawa2019distributionally}. 
It is based on the theorem that the representations of causal features elicit the existence of an optimal predictor across different domains.  
From a causal perspective, IRM identifies these causal features and excludes those features with spurious correlations as these correlations are not robust across different domains.
%
%
%
%
%
%
%
The connections between fairness and IRM are discussed in~\cite{arjovsky2019invariant,creager2020environment,veitch2021counterfactual}. IRM can learn representations to capture causal features which have invariant relationships to the prediction target. However, the representations may still contain the information of domains (e.g., different sensitive attributes), which may cause biases to prediction. Our work investigate to bridge this gap between IRM and counterfactual fairness.



\section{Conclusion}
In this work, we study a novel problem of learning counterfactually fair predictors from observational data with unknown causal models. We propose a principled framework \mymodel. More specifically, we specify this framework by learning counterfactually fair representations for each instance, and make predictions based on the representations. To learn fair representations, a variational auto-encoder based counterfactual data augmentation module is developed to generate counterfactual data with different values of sensitive attribute for each instance. We further reduce potential biases by applying the invariant penalty in each sensitive subgroup to exclude the 
variables with spurious correlations to the target. 
We evaluate the proposed framework under both real-world benchmark datasets and synthetic data. Extensive experimental results validate the superiority of the proposed framework over existing fairness predictors in different aspects. Overall, this paper provides insights for promoting counterfactual fairness in a more realistic scenario without given correct causal models, and also shows the impact of incorrect causal models. In the future, more research work on counterfactual fairness in real-world cases, such as missing and noisy data, is worth further exploration.

\section*{Acknowledgements}
Jing Ma, Aidong Zhang, and Jundong Li are supported by the National Science Foundation under grants (IIS-1955151, IIS-2006844, IIS-2008208, IIS-2106913, IIS-2144209, IIS-2223769, CNS-2154962, CNS-2213700, BCS-2228534, and CCF-2217071), the Commonwealth Cyber Initiative awards (VV-1Q23-007 and HV-2Q23-003), the JP Morgan Chase Faculty Research Award, the Cisco Faculty Research Award, the Jefferson Lab subcontract 23-D0163, the UVA 3 Cavaliers seed grant, and the 4-VA collaborative research grant.



\clearpage
\bibliographystyle{ACM-Reference-Format}
\balance
\bibliography{ref}
\clearpage
\appendix
\section{Implementation Details}
We use two fully connected layers in neural networks to implement $\Phi(\cdot)$, $g(\cdot)$ and $h(\cdot)$, respectively. The softmax function is used on top of $h(\cdot)$ when the sensitive attribute is categorical. LeakyRelu is used as activation functions in our framework. We aggregate the counterfactuals with mean operation, and we use mean square error (MSE) to compute the target prediction loss. For \mymodela, we adopt the implementation of MMD from~\cite{long2015learning}, and the optimization problem can be solved by traditional stochastic gradient descent algorithms. For \mymodelb, following \cite{chang2020invariant}, the minimax optimization problem is conducted with an alternating gradient descent process. We use cosine distance to implement $d(\cdot,\cdot)$. 

\section{Details of Experiment Settings}
\subsection{Full Introduction of Baselines}
\begin{itemize}
    \item \textbf{Constant Predictor:} A predictor which has constant output can obviously satisfy counterfactual fairness. We obtain this constant predictor by finding a constant which can minimize the mean squared error (MSE) loss on the training data.
    \item \textbf{Full Predictor:} Full predictor takes \textit{all} the observed attributes (except the attribute used as label) as input for prediction. We use linear regression for the regression task and logistic regression for the classification task. 
    \item \textbf{Unaware Predictor:} Unaware predictor is based on the notion of fairness through unawareness~\cite{grgic2016case}. It takes all features except the sensitive attribute as input to predict the label through linear regression for the regression task and logistic regression for the classification task.
    \item \textbf{Counterfactual Fairness Predictor:} We use two different counterfactual fairness predictors here: 1) As introduced in \cite{kusner2017counterfactual}, the predictor infers the latent variables and uses them along with the observed variables which are non-descendants of the sensitive attributes; 2) As described in \cite{russell2017worlds}, the predictor takes the input of both sensitive and non-sensitive attributes, with a fairness term added in the loss function which minimize the difference of the predictions made on two counterfactuals. We refer to these two methods as \cfpa~ and \cfpb, respectively. We follow the original implementations in \cite{kusner2017counterfactual,russell2017worlds}, where \cfpa~ uses linear regression for the regression task and logistic regression for the classification task, and \cfpb~ is implemented with neural networks.
\end{itemize}

\subsection{Detailed Experimental Setup}
We use Pyro \cite{bingham2019pyro} to implement the causal models. The number of sampling in the counterfactual generation is set as $500$. For the baselines CFP-U and CFP-O, the epochs for the causal model training is set as $2,000$ and the learning rate is set as $0.001$. All the presented results are averaged over ten executions of experiments. 

\section{More Experimental Results}
Table \ref{tab:imbalance_complete} shows the discrepancy of predictions made on different counterfactuals. In addition to the two pairs of counterfactuals ($S\leftarrow 0$ and $S\leftarrow 1$) and ($S\leftarrow 0$ and $S\leftarrow 2$) shown in Table \ref{tab:imbalance}, Table \ref{tab:imbalance_complete} also shows the results in pair ($S\leftarrow 2$ and $S\leftarrow 3$). Generally, the observation on the pair  ($S\leftarrow 2$ and $S\leftarrow 3$) is similar to the aforementioned observation on the pair ($S\leftarrow 0$ and $S\leftarrow 2$). 
\begin{table}[H]
\centering
 \caption{Study on synthetic data regarding the adverse effects of incorrect causal model $\mathcal{M}_2$.} 
 \vspace{-2mm}
 \label{tab:imbalance_complete}
  \begin{tabular}{l||cc}
    \hline
    \multirow{2}{*}{Method}  & \multicolumn{2}{c}{$S\leftarrow 2$ and $S\leftarrow 3$}\\ \cline{2-3}
     & MMD  & WASS  \\
    \hline
    \cfpa~(true)  & $8.407 \pm 0.810$  & $2.900\pm0.092$   \\
    \cfpa~(false) & $10.317 \pm 1.011$ & $3.780 \pm 0.052$ \\
    \cfpb~(true) & $8.793 \pm 0.927$  & $3.136 \pm 0.040$ \\
    \cfpb~(false) & $10.337 \pm 1.002$ & $3.864 \pm 0.030$ \\
    \hline
    \mymodela &  \underline{$8.108 \pm 0.024$}  & \underline{$2.860 \pm 0.004$} \\
    \mymodelb &  \bm{$7.902 \pm 0.055$}  & \bm{$2.761 \pm 0.005$}\\
    \hline
\end{tabular}
\end{table}




\end{document}